\documentclass[letterpaper, 10 pt, conference]{ieeeconf}  

\IEEEoverridecommandlockouts                              
\usepackage{epsfig} 
\usepackage{amsmath} 
\usepackage{amssymb}  
\usepackage{bm}

\def\N{{\mathcal N}}

\newcommand{\la}[1] {\label{#1} \iffalse  \quad{\rm{\bf #1 }}  \fi}
\newcommand{\be}{\begin{equation}}
\newcommand{\ee}{\end{equation}}
\newcommand{\beNN}{\begin{equation*}}
\newcommand{\eeNN}{\end{equation*}}
\newcommand{\ben}{\begin{enumerate}}
\newcommand{\een}{\end{enumerate}}
\newcommand{\bmat}{\left[ \begin{array}}
\newcommand{\emat}{\end{array} \right]}
\newcommand{\ba}{\begin{array}}
\newcommand{\ea}{\end{array}}
\newcommand{\bea}{\begin{eqnarray}}
\newcommand{\eea}{\end{eqnarray}}
\newcommand{\beaNN}{\begin{eqnarray*}}
\newcommand{\eeaNN}{\end{eqnarray*}}

\newcommand{\bi}{\begin{itemize}}
\newcommand{\ei}{\end{itemize}}
\newcommand{\bt}{\begin{tabular}}
\newcommand{\et}{\end{tabular}}

\title{\LARGE \bf Reconstruction error in a motion capture system}

\author{Andrea Masiero and Angelo Cenedese 
\thanks{A.Masiero A.Cenedese are with the Dipartimento di Ingegneria dell'Informazione,
        Universit\`a di Padova, via Gradenigo 6/B, 35131 Padova, Italy
        {\tt\small masiero@dei.unipd.it}, {\tt\small angelo.cenedese@unipd.it}.}
\thanks{The research leading to these results has received funding from the European Community's Seventh Framework Programme under agreement n.~FP7-ICT-223866 FeedNetBack and n.~257462 HYCON2 Network of excellence. This activity contributes to the seed project R3D of the Department of Information Engineering - University of Padova.}
}

\begin{document}

\maketitle
\thispagestyle{empty}
\pagestyle{empty}


\begin{abstract}
Marker-based motion capture (MoCap) systems can be composed by several dozens of cameras with the purpose of reconstructing the trajectories of hundreds of targets. With a large amount of cameras it becomes interesting to determine the optimal reconstruction strategy. For such aim it is of fundamental importance to understand the information provided by different camera measurements and how they are combined, i.e. how the reconstruction error changes by considering different cameras. In this work, first, an approximation of the reconstruction error variance is derived. The results obtained in some simulations suggest that the proposed strategy allows to obtain a good approximation of the real error variance with significant reduction of the computational time.
\end{abstract}


\section{Introduction}

Nowadays marker-based motion capture (MoCap) systems can be composed by several dozens of cameras with the purpose of reconstructing the trajectories of hundreds of targets. However, as costs of modern microprocessor and camera hardware decrease, it becomes economically viable to consider MoCap systems made of large camera networks of several hundreds of cameras, meeting the growing request for higher precision reconstruction of larger scenarios. This requirement, in terms of both minimizing the single target estimation error and of increasing the quality in the scene description, translates into scaling up with both the number of markers and the number of cameras.

The MoCap task can typically be divided into two steps: Reconstructing the 3D target positions by means of the measurements at time $t$, and merging such reconstructions with the dynamic evolutions of previously detected targets (data association and tracking). This paper focuses on the first step.
If the system is composed by a limited number of cameras and targets, the classical reconstruction algorithm based on geometric triangulation \cite{Hartley_1997,Hartley_book2003,Triggs_2000} can be implemented in a centralized fashion on a single machine to track the targets in real time. On the other hand, when considering the envisaged large system scenarios, it becomes difficult to simultaneously take into account the data provided by all the cameras. So, first, only portions of the system are considered simultaneously, and then the 3D reconstruction is achieved by progressively merging data from different parts of the system. In this framework it is important how the information is elaborated and merged by different cameras, i.e. some pairs of cameras will allow a better reconstruction\footnote{The concept of reconstruction quality considered here wants to take into account of several factors, among them: The number of reconstructed targets, the reconstruction accuracy, and the required computational time.} with respect to others.

This work deals with the problem of determining the information provided by different cameras about a target, and, consequently, what are the cameras that allow the optimal reconstruction of the investigated targets.

Even if the problem here is formulated in the MoCap framework, actually it is closely related also to other areas in computer vision: In the structure from motion framework \cite{MaSoatto_2004}, a quality measure among tensors is derived in \cite{Nister_2000} for reconstruction based on a hierarchical computational structure of trifocal tensors. Furthermore, in the multi-view stereo context, \cite{Goesele_2007} and \cite{Furukawa_2010} used suitable ``affinity'' functions to properly select a set of ``optimal'' views.


\section{Reconstruction error statistical description}
\label{Section:Gaussian_approximation}


Because of noise and discrete measurements the ray associated by a camera to a target's position actually represents only an estimation of the ``mean'' direction along which the target is positioned. The uncertainty on the 3D position provided by such mean direction grows as the distance between the viewing camera and the target increases (and it is parallel to the sensor plane, which practically is approximatively orthogonal to the ray direction).


%
%
%
%

The reconstruction error using two cameras changes depending on cameras' positions and orientations: Cameras at orthogonal orientations typically provides reconstructions with the smallest estimation errors, and conversely for cameras viewing along the same direction. However, because of the different views and of occlusions (both due to other objects in the scene and to the targets' object themselves), cameras at very different positions and orientations usually retrieve measurements of only few common targets.



In this section, the uncertainties on single and multi-camera reconstructions are presented in detail.


\subsection{Single camera measurements}

Consider a target $i$ placed at $\bar \phi_i=(\bar x_i, \bar y_i, \bar z_i)^\top$ in the 3D space. The noise of the target on the $j$-th camera measurement $\xi_{ij}=(u_{ij},v_{ij})^\top$ is assumed to be additive and Gaussian \cite{Hartley_1997,Hartley_book2003}:
\be
\xi_{ij} =  \bar \xi_{ij} + e_{ij}
~,
\ee
where $\bar \xi_{ij} = (\bar u_{ij},\bar v_{ij})^\top$ is the measurement without noise and $e_{ij} \sim \N(0,\Sigma_{e_{ij}})$ is the measurement noise. Hereafter the noise variance matrix $\Sigma_{e_{ij}}$ is modeled as $\text{diag}(\sigma_{e_{ij}}^2, \sigma_{e_{ij}}^2)$, where $\sigma_{e_{ij}}$ is the standard deviation.
Note that $\sigma_{e_{ij}}$ typically depends on camera and target reciprocal positions (and on camera orientation).
In addition, the value of $\sigma_{e_{ij}}$ depends also on the image analysis algorithm used for detecting it. Since the complete coverage of this topic is out of the scope of this work, hereafter the value of $\sigma_{e_{ij}}$ is taken as known.



\begin{figure}[ht!] \centering
\includegraphics[scale=0.5]{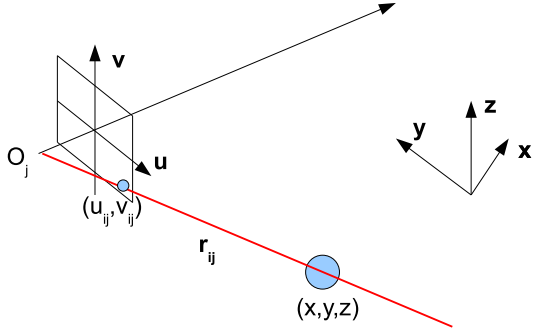}
\caption{Red line: ray associated to measurement $(u_{ij},v_{ij})$ of target $i$ in camera $j$.} \label{Fig:raggio}
\end{figure}

Each measurement from camera $j$ is a point on its sensor that corresponds to a ray passing through such point and camera's optical center, as shown in Fig.~\ref{Fig:raggio}.

\begin{figure}[ht!] \centering
\includegraphics[width=8.5cm]{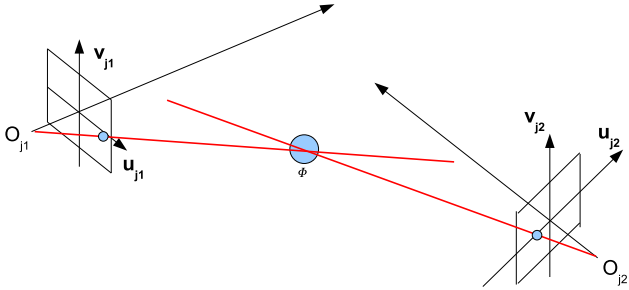}
\caption{Triangulation between two cameras. Crossing point between rays determined by different cameras allows to obtain target's 3D position.} \label{Fig:triangolazione}
\end{figure}

If different information are not available (e.g. the size of the target), target's 3D position cannot be reconstructed using a single measurement. However, it is possible to reconstruct by means of geometric triangulation using at least two measurements, as shown in Fig.~\ref{Fig:triangolazione}.

\begin{figure}[ht!] \centering
\includegraphics[width=8.5cm]{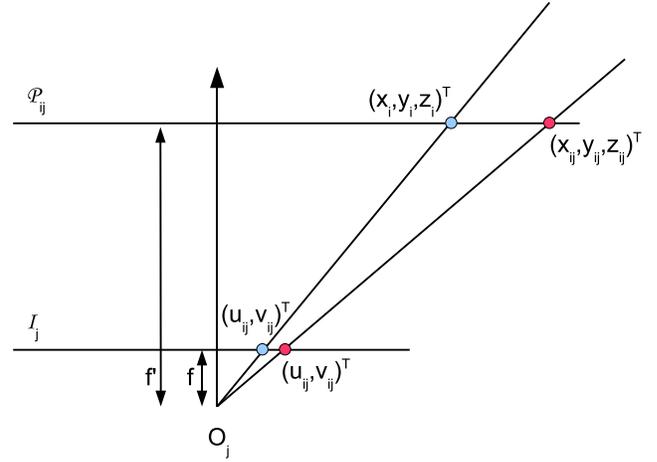}
\caption{Propagation of camera measurement error.} \label{Fig:errore1}
\end{figure}

Let the plane $\mathcal{P}_{ij}$ be parallel to the image plane $I_j$ of camera $j$ and passing through the target $i$. Because of the measurement noise $e_{ij}$ the ray associated to target $i$ by camera $j$ will intersect with plane $\mathcal{P}_{ij}$ on a point $\phi_i \neq \bar \phi_i$. Let $e'_{ij} = \phi_i - \bar \phi_i$: $e'_{ij}$ is obtained by propagation of error $e_{ij}$ according to:
\be
f e'_{ij} = f' e_{ij}
~,
\label{Eq:error_propagation}
\ee
where $f$ is the camera focal length and $f'$ is the distance from $I_j$ to $\mathcal{P}_{ij}$. Actually, the above equation holds for all planes $\mathcal{P}_j$ (on the front side of camera $j$) parallel to $I_j$ (see Fig.~\ref{Fig:errore1}).

While the measurement error propagates on the $\mathcal{P}_j$ plane as described, the measurement does not provide any information about the target position along the line starting from the optical center $O_j$ and passing through $\xi_{ij}$.

Exploiting different measurements of the same target allows to obtain a good estimation of the real distance $f'$, therefore by combining several camera measurements the information provided by camera $j$ about target $i$ can be modeled as:
\be
\phi_{ij} \sim \N \left( \bar \phi_{i} , \Sigma_{ij} \right)
~,
\label{Eq:single_meas_gaussian_approx}
\ee
where
\be
\Sigma_{ij} = M \psi_{ij} \psi_{ij}^\top + \sigma_{e_{ij}}^2 \left(\frac{f'}{f}\right)^2 \Psi_{ij} \Psi_{ij}^\top
~,
\label{Eq:var_single_meas_gaussian_approx}
\ee
and $M$ is a number much larger than the maximum room's side size multiplied by $m$, $\psi_{ij}$ is the unit vector along the direction from $O_j$ to $\bar \phi_i$, and $\Psi_{ij}$ is an orthonormal basis of the plane $P_{ij}$ parallel to $I_j$ and passing through the origin. Since $M$ is very large, the first term in \eqref{Eq:var_single_meas_gaussian_approx} expresses the practical absence of information provided by camera $j$ about target $i$ along the $\psi_{ij}$ direction, i.e. along the direction of the line from $O_j$ to the point.
Instead, the second term corresponds to the variance of the measurement error propagated using \eqref{Eq:error_propagation} to the plane $P_{ij}$. We stress the fact that the approximation of the reconstruction error variance \eqref{Eq:var_single_meas_gaussian_approx} is good in nonsingular conditions, i.e. when the target position can be adequately reconstructed (which is a typical operating condition when using a large number of cameras).
An experimental proof of the goodness of the approximation obtained by \eqref{Eq:var_single_meas_gaussian_approx} in the framework of multiple-cameras reconstruction is given by the simulations in the Subsec.~\ref{subsec:multiple}.


\subsection{Multiple camera reconstruction}
\label{subsec:multiple}

The approximation of equation \eqref{Eq:single_meas_gaussian_approx} is particularly useful when combining measurements from different cameras.
%

Without loss of generality, consider the reconstruction of the position of target $i$ from the measurements of cameras $j=1,2,\dots,m_i$. When at least two non aligned measurements are available, the position of the target can be estimated by means of geometric triangulation. Then, $f'$ in \eqref{Eq:error_propagation} is approximatively known, and \eqref{Eq:single_meas_gaussian_approx} is a good approximation of the information provided by each camera $j$ among those available for the reconstruction of target $i$. Thus, from \eqref{Eq:single_meas_gaussian_approx} the uncertainty on the reconstructed position can be approximated as follows (minimum variance estimation, \cite{Kailath_book2000}):

\be
\Sigma_{i} = \left( \sum_{j=1}^{m_i} \Sigma_{ij}^{-1} \right)^{-1}
~,
\label{Eq:variance_in_gaussian_approx_reconstruction}
\ee
and the overall standard deviation of the reconstruction error can be estimated as $\sqrt{ \text{trace}(\Sigma_i)}$.
%
%
%
%

\begin{figure}[ht!] \centering
\includegraphics[width=8.5cm]{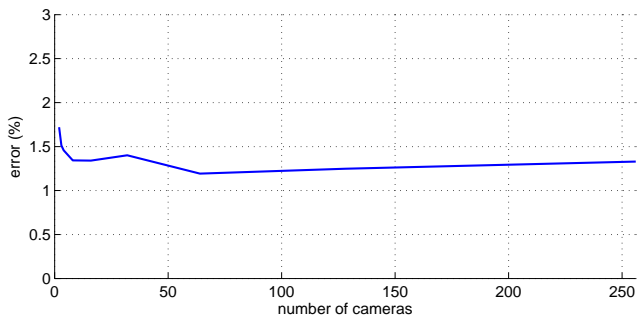}
\caption{Percent difference between the standard deviation computed from samples \eqref{Eq:MC_variance_estimation} and the approximated theoretical one \eqref{Eq:variance_in_gaussian_approx_reconstruction}. }
\label{Fig:sample_sd_vs_theor}
\end{figure}

For comparison, a direct evaluation of the reconstruction error variance can be obtained as the sample reconstruction variance in a Monte Carlo (MC) simulation:
\be
\hat \Sigma_i = \frac{1}{N-1} \sum_{k=1}^N  e_{i,k}e_{i,k}^\top
~,
\label{Eq:MC_variance_estimation}
\ee
where $N$ is a large integer number, and $e_{i,k}$ is the reconstruction error (difference between true and reconstructed position) in the MC iteration $k$.
Fig.~\ref{Fig:sample_sd_vs_theor} shows the percent error between the sample reconstruction standard deviation \eqref{Eq:MC_variance_estimation} and that computed from approximation \eqref{Eq:variance_in_gaussian_approx_reconstruction} varying the number of cameras $m$ considered for reconstruction from 2 to 256 (cameras are equally spaced along a circle of 10~m radius). The reported values are the mean of the results obtained on 1000 randomly sampled points (all positioned in the volume delimited by the cameras) for each choice of $m$. At each iteration, the $m$ cameras used for the reconstruction are randomly selected among the 256 available.


\begin{figure}[ht!] \centering
\includegraphics[width=8.5cm]{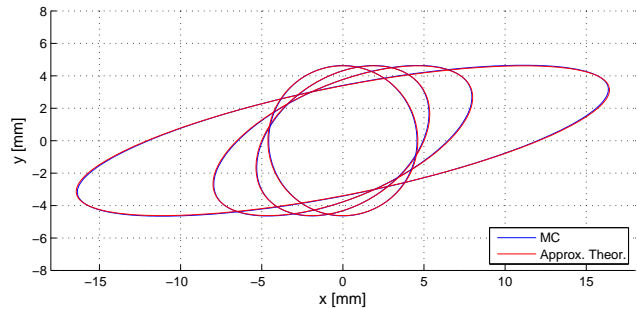}
\caption{Comparison of 1$\sigma$-level curve of the reconstruction error: Variance obtained by MC simulation (blue) and theoretical approximation (red). The error is evaluated for different angles between the two cameras: $\pi/8$ (which is represented by the external ellipse), $\pi/4$, $3\pi/8$ $\pi/2$ (small circle inside the other curves).}
\label{Fig:distribuzione_errore_ellissi}
\end{figure}

In Fig.~\ref{Fig:distribuzione_errore_ellissi}, it is highlighted how the reconstruction error (using two cameras) depends on the angle between the cameras: The closer the angle is to $\pi/2$ the better the estimated position results. In this example, 16 cameras are positioned (equally spaced) on a circle  of 10~m radius. The reconstruction error is computed for the point in the center of the cameras' circle. As shown in Fig.~\ref{Fig:distribuzione_errore_ellissi}, the 1$\sigma$-level curve computed from \eqref{Eq:variance_in_gaussian_approx_reconstruction} is practically overlapped to the 1$\sigma$-level curve estimated by sample data.


The performance evaluation of a MoCap system typically requires to compute the reconstruction error on a (quite large) representative number of points (voxels). Since the MC variance estimation can be quite time demanding, it is worth to consider \eqref{Eq:variance_in_gaussian_approx_reconstruction} that allows to compute in closed form good approximations (as in Figs.~\ref{Fig:sample_sd_vs_theor}-\ref{Fig:distribuzione_errore_ellissi}), at a computational cost largely lower than using the MC method.



\section{Conclusions}
\label{Sec:conclusions}

In this work, an approximation of the reconstruction error variance has been derived for marker-based motion capture system. Such approximation can be useful in deriving the optimal strategy for pairing cameras to reduce the reconstruction computational time in a distributed approach.


\bibliographystyle{plain}
\bibliography{mybib2}

\end{document}